# A SURVEY OF VOICE TRANSLATION METHODOLOGIES - ACOUSTIC DIALECT DECODER

Hans Krupakar
Dept. of Computer Science and Engineering
SSN College Of Engineering
E-mail: hans13033@cse.ssn.edu.in

Keerthika Rajvel
Dept. of Computer Science and Engineering
SSN College of Engineering
E-mail: keerthika13049@cse.ssn.edu.in

Bharathi B
Dept. of Computer Science and Engineering
SSN College Of Engineering
E-mail: bharathib@ssn.edu.in

Angel Deborah S
Dept. of Computer Science and Engineering
SSN College Of Engineering
E-mail: angeldeborahs@ssn.edu.in

Vallidevi Krishnamurthy
Dept. of Computer Science and Engineering
SSN College Of Engineering
E-mail: vallidevik@ssn.edu.in

**Abstract—** Speech Translation has always been about giving source text/audio input and waiting for system to give translated output in desired form. In this paper, we present the Acoustic Dialect Decoder (ADD) – a voice to voice ear-piece translation device. We introduce and survey the recent advances made in the field of Speech Engineering, to employ in the ADD, particularly focusing on the three major processing steps of Recognition, Translation and Synthesis. We tackle the problem of machine understanding of natural language by designing a recognition unit for source audio to text, a translation unit for source language text to target language text, and a synthesis unit for target language text to target language speech. Speech from the surroundings will be recorded by the recognition unit present on the ear-piece and translation will start as soon as one sentence is successfully read. This way, we hope to give translated output as and when input is being read. The recognition unit will use Hidden Markov Models (HMMs) Based Tool-Kit (HTK), hybrid RNN systems with gated memory cells, and the synthesis unit, HMM based speech synthesis system HTS. This system will initially be built as an English to Tamil translation device.

## 1. INTRODUCTION

Language is the one thing in the world that can both enable, and at the same time, completely shut out human communication. If it's a language known to us, we take hardly seconds to understand it. But if it's a language that we don't understand, it just cannot be understood without using dictionaries, manual parsers, translators and/or various applications available for translation. All of these solutions disrupt the flow of any conversation that someone could have with another person of a different dialect, because of the pause required to request for translation and time it takes for the actual translation process.

Automated simultaneous translation of natural language should have been a natural result in our multilingual world in order to make the process of communication amongst humans better, easier and efficient. However, the existing methods, including Google voice translators, typically handle the process of translation in a non-automated manner. This makes the process of translation of word(s) and/or sentences from one language to another, slower and more tedious. We wish to make that process automatic – have a device do what a human translator does, inside our ears.

### 1.1 MOTIVATION

Speech Translation is the process by which conversational spoken word(s) and/or phrase are translated and the result is obtained either in the form of words displayed on a screen or output being spoken aloud in the second language. The key to make this technology highly efficient is to automate this process and make it an inter-audio conversion so that it produces simultaneous results without having to physically start the process of translation. This enables people to simply wear the device and hear native speech in their own languages. When everyone in the world is equipped with one of these devices, there would be total understanding and harmony. This technology is of tremendous importance as it enables speakers of different languages to communicate and adds value to humankind in terms of World Peace, Science and Commerce, Cross-Cultural exchange, World Politics and Global Business.

## 2. RECOGNITION

Automatic speech recognition (ASR) can be defined as the independent, computer-driven transcription of spoken language into readable text in real time [1].

### 2.1 IMPLEMENTATION

Initially, the audio is input into the system. This audio is subjected to the process of Feature Extraction wherein noise and surrounding disturbances are removed to produce a feature vector. Grammar files for the input sentence are generated in Extended Backus Naur Form (EBNF). The vector is then trained using these grammar files to generate the audio input's corresponding textual sentence. This process is explained in a modular fashion in Figure 1.

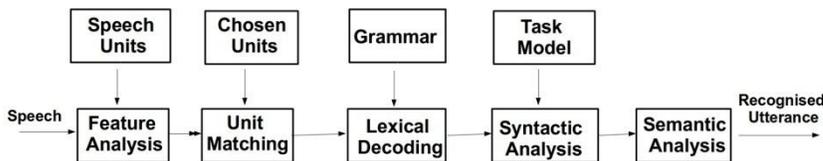

**FIG. 1 –SCHEMATIC OF HMM-BASED RECOGNITION** [2]

## 2.2 SURVEY OF AVAILABLE METHODS

There are a lot of methods used for ASR including HMMs, DTW, neural networks and deep neural networks[3]. HMMs are used in speech recognition because a speech signal can be visualised as a piecewise stationary signal or a short-time stationary signal. In a short time-scale (10ms), speech can be approximated as a stationary process. Speech can be thought of as a Markov model for many stochastic purposes. Another reason why HMMs are popular is because they can be trained automatically and are simple and computationally feasible to use[4]. Dynamic time warping was used before HMMs in SR but has long been declared less successful and isn't used[5]. Neural networks can be used efficiently in SR but are rarely successful for continuous recognition tasks because of their lack of ability to model temporal dependencies [6].Deep neural networks have been shown to succeed in SR but because it is a system of 3 components, recognition, translation and synthesis, we have used HMMs with the least time complexity.

SR can be further divided into word-based, phrase-based and phoneme-based mechanisms. While there is excellent performance in isolated word recognition with word-based models, they fail when it comes to continuous SR because of complexity. Phoneme-based is the best approach and is employed in this system because of its ability to incorporate and work with large corpus and ease of addition of new words into the vocabulary [7].

## 2.3 SURVEY OF FEATURE EXTRACTION METHODS

The main component of ASR is Feature Extraction. It uses maximum relevant information about the speech signal and helps distinguish between different linguistic units and removes external noise, disturbances and emotions [2].

Commonly used feature extraction techniques are MFCCs (Mel Frequency Cepstral Coefficients), LPCs (Linear Prediction Coefficients) and LPCCs (Linear Prediction Cepstral Coefficients) [8]. MFCCs feature shall be used for this process and the reason is stated as follows. MFCCs have been ascertained as the state-of-the-art for feature extraction especially since it is based on actual human auditory system and has a perceptual frequency scale called Mel - frequency scale. It combines the advantage of the Cepstral analysis with a perceptual frequency scale based on critical bands making use of logarithmically spaced filter banks [9]. Prior to the introduction of MFCCs, LPCs and LPCCs were used for the feature extraction process. These have been stated as obsolete since both have a linear computation nature and lack perceptual frequency scales even though they are efficient in tuning out environmental noise and other disturbances from the sampled speech signal [10]. Further, several researches based on feature extraction for various languages like English, Chinese, Japanese and even Indian languages like Hindi have proved experimentally that MFCCs produce at least 80 percent efficiency as opposed to just 60 percent by LPCs and LPCCs [11].

## 2.4 SURVEY OF TOOLS FOR ASR

There are numerous tools developed for ASR such as HTK, JULIUS, SPHINX, KALDI, iATROS, AUDACITY, PRAAT, SCARF, RWTH ASR and CSL.

Out of all the available, the 2 fairly popular and widely accepted frameworks are HTK and SPHINX. Both of these are based on Hidden Markov Model(HMM) and are open source. Both frameworks can be used to develop, train, and test a speech model from existing corpus speech utterance data by using Hidden Markov modeling techniques [12]. Figure 2 shows the results that were achieved decoding the test data from AN4 corpus. These were achieved on a PC running Windows and Cygwin, with a 2.6GHz Pentium 4 processor with 2 GB System RAM [12].

| Metric | Sphinx3 | HTK |
| --- | --- | --- |
| Peak Memory Usage (MB) | 8.2 | 5.9 |
| Time to Completion (sec) | 63 | 93 |
| Sentence Error Rate (%) | 59.2 | 69.0 |
| Word Error Rate (%) | 21.3 | 9.0 |
| Word Substitution Errors | 92 | 92 |
| Word Insertion Errors | 71 | 154 |
| Word Deletion Errors | 2 | 0 |

**FIG. 2 – RESULTS FROM AN4 CORPUS**

HTK decoder did not make any deletions, which gave it a slight advantage on the overall word error rate. Also, while HVite uses less memory during decoding, the time difference in running the test set is significant at 30 seconds [12], [13]. In addition to this HTK supports HCopy tools which provide a wealth of input/output combinations of model data and front-end features couple of which are not present in SPHINX but certain compatibility is provided. However, the efficiency of compatibility is debatable [14]. Other slight advantages of HTK over SPHINX are that HTK is supported on multiple OS while SPHINX is largely supported on LINUX platforms only. Also, HTK uses C

Programming language while SPHINX uses the Java Programming language [15].

## 3. TRANSLATION

Machine Translation (MT) is the process of converting a source sentence sequence into the target sentence sequence of same/different length. Even though MT has come a long way from where it was with its initial models, it is nowhere near being completely efficient. Machine translation has been worked on for decades now and the recent advancements of using neural networks have propelled the field to a new height. MT can be direct, rule-based or data-driven.

### 3.1 HISTORY

Machine translation has been implemented in multitudes of methodologies ranging over decades of research. Machine Translation saw its advent with the direct translation system, and then more ways were introduced like rule-based and Example-Based Machine Translation (EBMT) systems. Direct translation is the process conditional transcription of source and target words. Rule-based systems operated on rules to translate sentences while the example-based system mapped via examples. The problems with these approaches are the lack of interlingua database and scalability, naive nature of translations etc [15].

Statistical Machine Translation is one where the source sentence is encoded into a representation which is translated into target language by maximising the probability of the closeness of the target sentence by using Bayes rule. It is faster than any of the systems used before it because of its parallel processing of various modules or subsystems. It is not scalable to large scale MT because of memory and performance bottlenecks [16].

### 3.2 RNNs IN MT - ENCODER DECODER APPROACH

While SMT is better than the previously used MT techniques, there was still an inherent complexity issue. More importantly, SMT systems took up a lot of space even for a translation engine with very small vocabularies. The advent of Recurrent Neural Networks (RNNs) into Natural Language Processing saw the field take a turn to an unexpected advancement. RNNs can be defined by means of:

$$h_{<t>} = f(h_{<t-1>}, x_{<t>})$$

$$y_{<t>} = g(h_{<t>})$$

where f is a smooth bounded function, $h_t$ refers to the hidden layer function at time epoch t, $x_t$ refers to the input received by RNN at time t, $y_t$ the output at time t, and g known as the output function. Usually, the input to RNNs is encoded in the hidden layer using functions that map the input onto the continuous space like tanh, sigmoid function etc. The output function is an activation function and the most common ones are the SoftMax activation function, sigmoid functions, tanh etc [17].

The first RNN systems used in MT involved one RNN that takes in the source sentence as input, word by word, mutating the input using the hidden layer where the output function mapped the encoded form of the input from the hidden layer back into a form that can be used to arrive at the target sentence using language and translation modelling functions. In the beginning, there was only one hidden layer used and this type of RNNs are known as shallow RNNs. Later, more layers of hidden functions were used instead of just one or two and this system is known as Deep RNN (DRNNs). DRNNs were found to be much better at encoding the source sentence into the target sentence and thus quickly replaced shallow RNNs in MT. The approach of using a single RNN for MT is widely known as the phrase-based approach [18].

One of the more recent methods in MT is the encoder-decoder approach. This method, shown in figure 3, involves two RNNs, one for encoding and the other for decoding. The encoder RNN takes the source sentence, word by word, and transforms it into a vector that contains all the properties of the sentence as the hidden layer function is recursive on its previous function. The decoder RNN then takes this vector and maps it to the target sentence [19], [20].

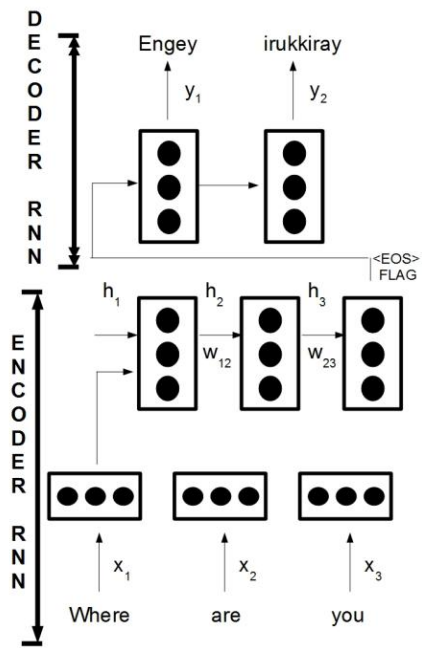

**FIG. 3 – ENCODER DECODER APPROACH** [19]

While this system, theoretically, is perfect in retaining long range dependencies across the input sentence, in practice, it falls short because the system is unable to maintain these dependencies without memory units. This made it impossible for this system to translate sentences of a bigger sentence length. This is popularly known as the vanishing gradient problem. To get rid of this problem, there have been various methods implemented [20].

## 3.3 AUTOMATIC SEGMENTATION

Automatic Segmentation was one of the first such methods where meaningful phrases were translated together after successful segmentation. The issue of sentence length occurs because the Neural Network fails to recognize some of the initial words of the input sentence from the vector that the sentence is transcribed to. The input sentence is segmented into cohesive phrases that can be translated easily by the NN. After every segment of the source sentence is translated into the target sentence, the target phrases are concatenated together to produce output.

While this gets rid of the vanishing gradient problem, it poses a few new difficulties. Because the neural network can only work with easy to translate (cohesive) clauses, there is a clause division problem because the system cannot decide the best cohesive phrases to translate into target language easily. Also, computational complexity increases because of the parallel processing required in reading input words and translating previously read phrase at the same time. Also, this method of concatenating translated phrases only works with languages that don't have long range dependencies between the words of a sentence and only with source target language pairs with semantically similar grammar structures.[21].

## 3.4 BIDIRECTIONAL RNNs:

One of the main problems of RNNs is the inability of the system to maintain long range dependencies for sentences with a lot of words. One of the main reasons for this is that the input sequence is only scanned in one direction, normally from the beginning of the sentence to the end. To simultaneously model both past and future references, bidirectional RNNs should be used [22]. The RNN system is composed of two independent recurrent layers: one layer processes the input sentence in forward time steps 1 to T (first word through to last word), while the other layer processes the input sentence in backward time steps from T to 1 (last word to first word). Bidirectional RNNs are defined by an output function y, 2 hidden states, $h^f$ for forward time steps and $h^r$ for the backward time steps as:

$$h^f_{<t>} = f_h(h^f_{<t-1>}, x_{<t>})$$

$$h^r_{<t>} = f_h(h^r_{<t-1>}, x_{<t>})$$

$$y_{<t>} = f_o(h^f_{<t-1>}, h^r_{<t-1>}, x_{<t>})$$

The hidden state functions can be a simple sigmoid function or a complex LSTM network [7]. The use of BRNNs in phrase-based SMT is implemented with the help of n-best lists as the systems are complementary. While the translation quality was significantly better than using unidirectional RNNs, this particular BRNN MT system did not come close to the current best translation quality.

BRNNs were also later used in many Encoder Decoder models, one of which had a Bi-Directional Decoder RNN used to model each word to summarize both the preceding and succeeding words. This method aligned and simultaneously translated input [23], [24].

## 3.5 LSTM

Because regular RNN hidden layers are unable to successfully store information about the sentence words in them, the hidden layers are built along with a gate-operated memory unit that is capable of retaining the encoding done in the state for a long time. This solves the problem of lack of long range dependencies in the system. To ensure that the system doesn't have errors, an algorithm like back propagation, Stochastic Gradient Descent (SGD), Linear Gradient Descent etc. is used to normalise the values. The main problem with regular RNN systems is that it doesn't retain the values for these algorithms to be applied. An LSTM neuron is defined as:

$$i_{<t>} = \xi(W_{xi}x_{<t>} + W_{hi}h_{<t-1>} + W_{ci}c_{<t-1>} + b_i)$$

$$f_{<t>} = \xi(W_{xf}x_{<t>} + W_{hf}h_{<t-1>} + W_{cf}c_{<t-1>} + b_f)$$

$$c_{<t>} = f_{<t>}c_{<t-1>} + i_{<t>}\tanh(W_{xc}x_{<t>} + W_{hc}h_{<t-1>} + b_c)$$

$$o_{<t>} = \xi(W_{xo}x_{<t>} + W_{ho}h_{<t-1>} + W_{co}c_{<t-1>} + b_o)$$

$$h_{<t>} = o_{<t>}\tanh(c_{<t>})$$

$$y_{<t>} = g(h_{<t>})$$

where i, f, c, and o are the input, forget, cell and output gates respectively. The hidden layer depends on the output and the cell gates. The output of each hidden layer is a function of the hidden layer at that time epoch t. $\xi$ is the logistic sigmoid function and $w_{ij}$ refers to the weight of the edge from i to j gates. This is explained in figure 4 [25], [26].

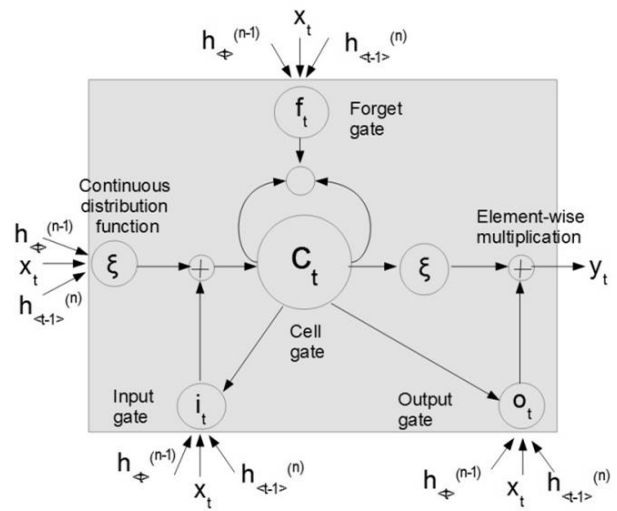

**FIG. 4 – LSTM NETWORK SCHEMA** [28]

There were significant improvements in performance of the system when compared to traditional RNNs. Interestingly, it was found that the performance increased when the sentence was input in reverse order because of the structural similarities in the languages of English and French that this system was implemented in [27].

### 3.6 GRU

Gated Recurrent Units (GRUs) are a variation of the well known LSTM approach where the neuron has gated mechanisms that enable it to remember encodings like LSTMs but unlike the LSTM, it has no memory unit of its own. GRUs, shown in figure 5, are defined by:

$$h_{<t>} = (1 - z_{<t>})h_{<t-1>} + z_{<t>}\bar{h}_{<t>}$$

$$z_{<t>} = \xi(W_{xt} + U_{zh} - 1)$$

$$\bar{h}_{<t>} = tanh(W_{xt} + U(r_{<t>} \odot h_{<t-1>}))$$

$$r_{<t>} = \xi(W_r x_{<t>} + U_r h_{<t-1>})$$

$$y_{<t>} = g(h_{<t>})$$

where $\odot$ is element wise multiplication operator, y is the output function, h is the hidden state function, z is the update gate, r is the reset gate and $\bar{h}$ is the candidate activation function of the Gated Recurrent Unit, $\xi$ is the sigmoid function.

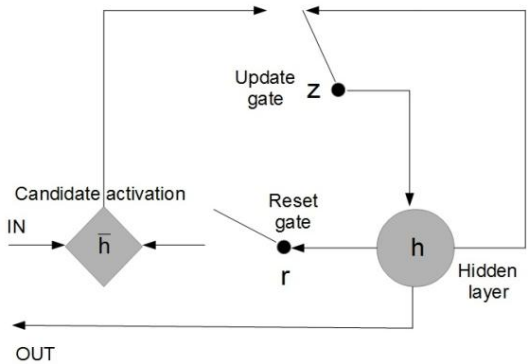

**FIG. 5 – GRU NETWORK SCHEMA** [29]

GRUs have proven to be really close in terms of performance with the LSTM networks and there is really nothing that can say that one is clearly one that is better than the other. Both systems have been interchangeably used in MT with similar resulting translations[30].

### 3.7 LARGE VOCABULARY PROBLEM:

The methods of machine translation described by various RNN approaches fail to acknowledge the problem of a target vocabulary. The presence of a very big target vocabulary makes it computationally infeasible if the number of words in the target language known to the system exceeds a threshold amount. To address this problem, a hybrid encoder decoder system with an attention mechanism has been used [23]. The algorithm proposed successfully manages to keep the computational complexity to only a part of the vocabulary.

In the system proposed, a Bi-Directional RNN with a Gated Recurrent Unit (GRU) is used as the encoder to ensure that the encoding process is very efficient and at the same time, faster than conventional BRNNs because the gated unit skips over time epochs. The decoder computes the context vector $c_{<t>}$ as a convex sum of the hidden states ($h_1, \ldots, h_T$) with the coefficients $\alpha_1, \ldots, \alpha_T$ computed by

$$a_{<t>} = \frac{\exp(a(h_{<t>}, z_{<t-1>}))}{\sum_k \exp(a(h_{<k>}, z_{<t-1>}))}$$

where a is a feed forward NN with a single hidden layer z.

One of the major hurdles in this system is the scaling of complexity of accessing a large target vocabulary. There are 2 approaches to deal with the problem of complexity due to a large vocabulary:

Stochastic Approximation of Probability: In this method, the target word is best estimated using a noise contrastive estimation.

Hierarchy Classes: In this approach, the target words are clustered hierarchically into multiple classes such that the target probability is based on the class probability and the inter-class probability.

The Rare words model is a translation-specific solution for this problem. In this approach, only a small subset of the target vocabulary is used to compute the normalization constant during training, making complexity constant with respect to the target vocabulary. Also, after each update, the complexity is brought down [31].

The easiest way to select a part of the vocabulary is to select top N most frequent target words but this would ruin the point of having a large vocabulary. This model creates a dictionary based on source and target word alignment. Using this dictionary, K best choices are chosen and this is further scrutinized for the final output. Also, it obtained a BLEU score of 37.2 for English to French translation which is just 0.3 behind the current best. Also, it was able to perform really efficiently even though it had a large target vocabulary[32].

### 4. SYNTHESIS

Speech synthesis is the process of generating computer simulation of human speech. It is used to translate written information/text into aural information. It has been the counterpart of speech recognition.

## 4.1 IMPLEMENTATION

The textual input sentence is first pre-processed by a process called Normalization where things like special characters, date and time, numbers, and abbreviations are turned into words. Next, a list of phonemes, taken from the list of database of phonemes used in the language, is used for prosody generation of the speech sounds of the target speech sentences [33]. The phonemes are chosen based on the spectral parameter, the intonation based on the excitation parameter, and the duration, based on the duration parameter. This system is shown in figure 6.

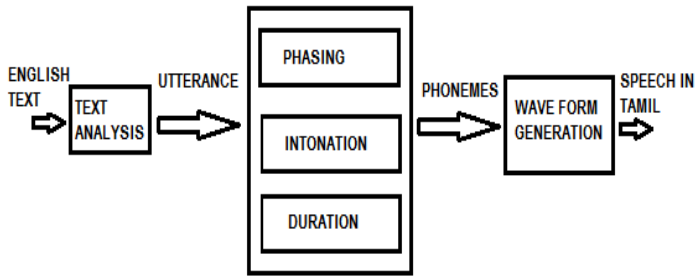

**FIG. 6 – SCHEMATIC OF HMM-BASED SYNTHESIS**

## 4.2 IMPORTANCE OF PROSODY [33], [34]

The most important qualities of a speech synthesis system are naturalness- how closely output sounds like human speech and intelligibility- the ease with which the output is understood. Appropriate Prosody model is essential to ensure naturalness and intelligibility as it serves as the backbone of TTS system. Prosody means the characteristics that are obtained from the speech like accent, intonation and rhythm. These parameters have information of duration, pitch and intensity. Earlier rule-based approach was used for deriving the prosody modelling for concatenative synthesis. Today, statistical approaches are popularly adopted. Also cues are provided by the prosody to the listener to help them interpret the speech correctly. Factors like way of speaking, regional effect and various other phonological factors affect the prosody.

## 4.3 SYNTHESIS TECHNOLOGIES

There are three main approaches to speech synthesis: Formant synthesis, Articulatory synthesis, and Concatenative synthesis.

### 4.3.1 FORMANT SYNTHESIS

Formants are a set of resonance frequencies of the vocal tract. Formant synthesis models the frequencies of speech signal. It does not use human speech samples instead creates an artificial speech using parameters such as fundamental frequency, voicing, and noise levels are varied over time to create a waveform of artificial speech [35]. It results in robotic sounding speech.

### 4.3.2 ARTICULATORY SYNTHESIS

Articulatory synthesis tries to model the human speech production system and articulatory processes directly. However, it is the most difficult method to implement due to lack of knowledge of the complex human articulation organs[35].

### 4.3.3 CONCATENATIVE SYNTHESIS

Concatenative synthesis is based on the concatenation of segments of recorded speech. It produces the most natural sounding synthesized speech. However, it has serious drawbacks like audible glitches in the output sometimes and the memory requirement is large to store a large amount of speech corpus [36].

### 4.3.4 HMM BASED SYNTHESIS

It is a statistical parametric synthesis technique. It is used easily for implementing prosody and various voice characteristics on the basis of probabilities without having large databases. In this system, the frequency spectrum, fundamental frequency, and prosody of speech are modeled simultaneously by HMMs [37]. Speech waveforms are generated from HMMs on the basis of maximum likelihood criterion. In this approach speech utterances are used to extract spectral (Mel-Cepstral Coefficients.), excitation parameters and model context dependent phone models which are, in turn, concatenated and used to synthesize speech waveform corresponding to the text input [35], [36].HTS technology is preferred because it overcomes the drawbacks of Formant and Articulatory synthesis as HMM based is a statistical approach.

## 5. USE OF 3 INDEPENDENT MODULES:

Spontaneous speech poses a very important problem in the process of translation. This occurs mainly because the variation is speech patterns, accents, intonations etc. make it impossible to detect even the right sentence let alone the errors present in the 3 modules employed. Most of these errors are caused by the lack of accuracy of the recognition. As a result, the input sentence is not well formed. Even without recognition errors, speech translation cannot rely on conventional grammar models and structures because they differ from those of written language because of the nature of speech.

Recently, SMT has shown promise in voice translation. SMTs don't have the need to make syntactic assumptions because of the statistical nature of the system. A target sentence is guaranteed to be output by the system regardless of the nature of input. This ensures that even if there is no syntactic and structural accuracy in translation, at least the same meaning is retained in the translation. Having said that, the SMT structure of recognition followed by translation and synthesis lacks coherent a working style because of the very independent approach of each of these modules. Also, there are a lot of models like n-best lists, n-gram model, bag of words model etc. that can be used along with the SMT system to increase the performance of the system quite drastically [38], [39].

## 6. PROPOSED SYSTEM

The proposed system is composed of three processes- recognition, translation and synthesis. Fig. 1 describes the processes involved. Section 2 begins by describing the working of HMM-Based Recognition and supports the decision for choosing that methodology by surveying the other tools and methods available to perform SR. Section 3 contains the survey of the various translation methodologies and with focus on the adaptation of LSTM and GRU, and then addressing the target vocabulary problem. This helps make the choice to use gated memory networks and try out several variations to see optimality in output. Section 4 explains the HMM-Based synthesis process and supports the decision with survey of the other available methodologies, thus making it the perfect methodology for the task. Further, Section 5 mentions the drawbacks of using three independent modules in voice translation. It also talks about complementary language models that can be used to enhance performance. The survey concludes by stating the requirements and expectations of the model. The scope for extension of the services provided by ADD is described in Section 8.

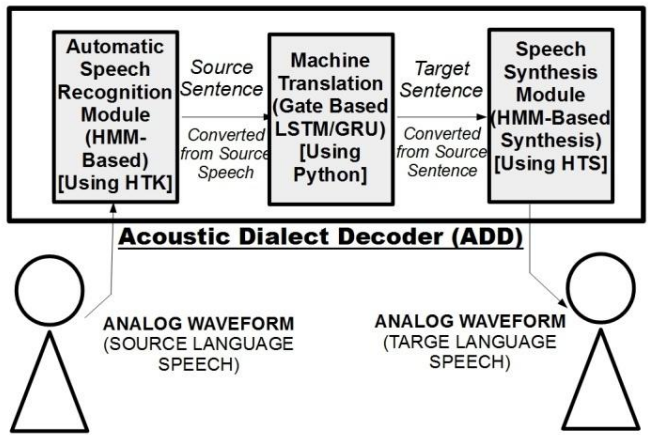

**FIG. 7 - THE PROPOSED SYSTEM**

This Acoustic Dialect Decoder is a project undertaking funded by SSN College Of Engineering for Rs. 20,000.

## 7. CONCLUSION

We have found that HMM based Speech Recognition, hybrid RNN based Machine Translation based on gated units and other approaches, and HMM based Speech Synthesis to be the best approaches in the respective paradigms. Therefore, the objective is to build a continuous speaker-independent English - Tamil Voice Translation system using HMM based speech recognition in HTK, hybrid RNN system with at least one LSTM or GRU based Machine Translation system in Python, and HMM based speech synthesis on HTS. The idea is to work with medium to fairly large vocabulary size and improve efficiency and accuracy of the system.

## 8. FUTURE WORK

ADD has huge scope for improvement and extension. It can be made to translate multiple languages using the same input structures. It can be made much more efficient using various efficiency enhancing algorithms like the rare words model and many others. It can also be improved to make the target speech sound exactly like the source speaker to enhance the comfort of using the device. This system can also be incorporated into a hearing aid to enable the same service for people who are deaf as well.